\documentclass[pmlr]{jmlr}

\usepackage{longtable}

\usepackage{booktabs}
\usepackage[load-configurations=version-1]{siunitx} 

\makeatletter
\def\set@curr@file#1{\def\@curr@file{#1}} 
\makeatother

\theorembodyfont{\upshape}
\theoremheaderfont{\scshape}
\theorempostheader{:}
\theoremsep{\newline}

\usepackage{enumitem}

\title[CERT: Contrastive Self-supervised Learning for Language Understanding]{CERT: Contrastive Self-supervised Learning for Language Understanding}

\author{\Name{Hongchao Fang\textsuperscript{$\dagger$}}
       \Email{colefang.hongchao@gmail.com} \\
       \addr UC San Diego
       \AND
\Name{Sicheng Wang\textsuperscript{$\dagger$*}}
       \Email{wsc879453141@gmail.com} \\
       \addr UC San Diego 
\AND
 \Name{Meng Zhou\textsuperscript{$\dagger$*}}
       \Email{zhoumeng9904@sjtu.edu.cn}\\
       \addr UC San Diego
       \AND
        \Name{Jiayuan Ding\textsuperscript{*}}
       \Email{jiayuand@usc.edu}\\
       \addr VMware 
       \AND
       \Name{Pengtao Xie}
       \Email{pengtaoxie2008@gmail.com}\\
       \addr UC San Diego
\AND
       }

\begin{document}

\maketitle

\begin{abstract}
Pretrained language models such as BERT, GPT have shown great effectiveness in language understanding. The auxiliary predictive tasks in existing pretraining approaches are mostly defined on tokens, thus may not be able to capture sentence-level semantics very well. To address this issue,  we propose CERT: Contrastive self-supervised Encoder Representations from Transformers, which pretrains language representation models using contrastive self-supervised learning at the sentence level. CERT creates augmentations of original sentences using back-translation. Then it finetunes a pretrained language encoder (e.g., BERT) by predicting whether two augmented sentences originate from the same sentence. CERT is simple to use and can be flexibly plugged into any pretraining-finetuning NLP pipeline. We evaluate CERT on 11 natural language understanding tasks in the GLUE benchmark where CERT outperforms BERT on 7 tasks, achieves the same performance as BERT on 2 tasks, and performs worse than BERT on 2 tasks. On the averaged score of the 11 tasks, CERT outperforms BERT.  The data and code are available at \url{https://github.com/UCSD-AI4H/CERT}
\end{abstract}

\section{Introduction}

\let\thefootnote\relax\footnotetext{$\dagger$The work was done during internship at UCSD.}
\let\thefootnote\relax\footnotetext{$*$Equal contribution.}

Large-scale pretrained language representation models such as BERT~\citep{devlin2018bert}, GPT~\citep{radford2018improving}, BART~\citep{lewis2019bart}, etc. have achieved dominating performance in various natural language processing tasks, such as text generation, reading comprehension, text classification, etc. The architectures of these models are mostly based on Transformer~\citep{vaswani2017attention}, which uses self-attention to capture long-range dependency between tokens. The Transformer encoder or decoder is pretrained on large-scale text corpus by solving self-supervised tasks, such as predicting masked tokens~\citep{devlin2018bert}, generating future tokens~\citep{radford2018improving}, denoising corrupted tokens~\citep{lewis2019bart}, etc. In these works, the targets to be predicted are mostly  at the word level. As a result, the global semantics at the sentence level may not be sufficiently captured.

To address this issue, we propose CERT: Contrastive self-supervised Encoder Representations from Transformers, which uses contrastive self-supervised learning (CSSL)~\citep{he2019moco,chen2020simple} to learn sentence-level representations. Recently, CSSL has shown promising results in learning visual representations in an unsupervised way~\citep{he2019moco,chen2020simple}. The key idea of CSSL is: create augments of original examples, then learn representations by predicting whether two augments are from the same original data example or not. CERT creates augments of sentences using back-translation~\citep{edunov2018understanding}, then finetunes a pretrained language representation model (e.g., BERT, BART) by predicting whether two augments are from the same original sentence or not. Different from existing pretraining methods where the prediction tasks are defined on tokens, CERT defines the prediction task on sentences, which can presumably better capture global semantics at the sentence level. 

CERT uses back-translation~\citep{edunov2018understanding} to perform sentence augmentation. Given a sentence $x$ in source language $S$, we use an S-to-T translation model to translate $x$ into a sentence $y$ in target language $T$. Then we use a T-to-S translation model to translate $y$ into $x'$ in the source language. $x'$ is regarded as an augment of $x$. Translation models for different target languages are used to create different augments of a source sentence. Given these augmented sentences, a Momentum Contrast (MoCo)~\citep{he2019moco} method is used to perform CSSL. MoCo maintains a queue of augmented sentences (called keys) which are encoded using a pretrained text-encoder (e.g., BERT) with  momentum updates. Given an augmented sentence (called query), a similarity score is calculated between the BERT (or any other pretrained text-encoder)  encoding of the query and each key in the queue. The query and a key are labeled as a positive pair if they are augments of the same original sentence and as a negative pair if otherwise. These binary labels and similarity scores are used to define contrastive losses~\citep{hadsell2006dimensionality}. The weights of the pretrained text encoder are further pretrained by minimizing the contrastive losses. To apply the pretrained CERT model on a downstream task, we finetune the CERT weights using input data and labels from the downstream task. 
CERT is a flexible module that can be integrated with any pretrained language representation models, such as BERT, BART, ERNIE 2.0~\citep{sun2019ernie}, T5~\citep{raffel2019exploring}, etc. We evaluate CERT on 11 natural language understanding tasks in the GLUE~\citep{wang2018glue} benchmark where CERT outperforms BERT on 7 tasks, achieves the same performance as BERT on 2 tasks, and performs worse than BERT on 2 tasks. On the averaged score of the 11 tasks, CERT outperforms BERT. These results demonstrate  the effectiveness of contrastive self-supervised learning for language representation by capturing sentence-level semantics. 



The major contributions of this paper are as follows:
\begin{itemize}[leftmargin=*]
\item We propose CERT, a new language representation pretraining method based on contrastive self-supervised learning. The predictive tasks of CERT are defined at the sentence level, thus presumably can better capture sentence-level semantics.
\item We perform evaluation of CERT on  11  natural language understanding tasks in the GLUE benchmark, where CERT outperforms BERT on the averaged score of 11 tasks.
\item We perform ablation studies to investigate how the performance of CERT is affected by sentence augmentation methods and the source of pretraining corpora.
\end{itemize}

The rest of the papers are organized as follows. Section 2 and 3 present the methods and experiments. Section 4 reviews related works and Section 5 concludes the paper.

\section{Pretraining of Transformers for Language Understanding}
Among the recent works for pretraining language representation models, most of them are based on the Transformer~\citep{vaswani2017attention} architecture. For example, BERT pretrains Transformer encoder. GPT pretrains  Transformer decoder. BART pretrains Transformer encoder and decoder jointly. 

\subsection{Transformer}

 Transformer~\citep{vaswani2017attention} is an encode-decoder architecture for sequence-to-sequence (seq2seq)  modeling~\citep{sutskever2014sequence}. Different from seq2seq models~\citep{sutskever2014sequence} that are based on recurrent neural networks (e.g., LSTM~\citep{hochreiter1997long}, GRU~\citep{chung2014empirical}) which model a sequence of tokens via a recurrent manner and hence is computationally inefficient. Transformer eschews recurrent computation and instead uses self-attention which not only can capture the dependency between tokens but also is amenable for parallel computation with high efficiency. Self-attention calculates the correlation among every pair of tokens and uses these correlation scores to create ``attentive" representations by taking weighted summation of tokens' embeddings. Transformer is composed of building blocks, each consisting of a self-attention layer and a position-wise feed-forward layer.  Residual connection \citep{he2016deep} is applied around each of the two sub-layers, followed by layer normalization~\citep{ba2016layer}. Given the input sequence, an encoder, which is a stack of such building blocks, is applied to obtain a representation for each token. Then the decoder takes these representations as inputs and decodes the sequence of output tokens. To decode the $i$-th token, the decoder first uses self-attention to encode the already decoded sequence $y_1,\cdots,y_{i-1}$, then performs input-output attention between the encodings of $y_1,\cdots,y_{i-1}$ and those of the input sequence. The ``attentive" representations are then fed into a feed-forward layer. The three steps are repeated for multiple times. Finally, the representation is fed into a linear layer to predict the next token. The weight parameters in Transformer are learned by maximizing the conditional likelihood of output sequences conditioned on the corresponding input sequences.

\subsection{BERT}

BERT~\citep{devlin2018bert} aims to  learn a Transformer encoder for representing texts. BERT’s model architecture is a multi-layer bidirectional Transformer encoder. In BERT, the Transformer uses bidirectional self-attention. To train the encoder, BERT masks some percentage of the input
tokens at random, and then predicts those masked tokens by feeding the final hidden vectors (produced by the encoder) corresponding to the mask tokens into an output softmax over
the vocabulary. To apply the pretrained BERT to a downstream task such as sentence classification, one can add an additional layer on top of the BERT architecture and train this newly-added layer using the labeled data in the target task.

\section{Contrastive Self-supervised Learning}
Self-supervised learning (SSL)~\citep{wu2018unsupervised,he2019moco,chen2020mocov2,chen2020simple} is a learning paradigm which aims to capture the intrinsic patterns and properties of input data  without using human-provided labels. 
The basic idea of SSL is to construct some auxiliary tasks solely based on the input data itself without using human-annotated labels and force the network to learn meaningful representations by performing the auxiliary tasks well, such rotation prediction~\citep{gidaris2018unsupervised},  image inpainting~\citep{pathak2016context},  automatic colorization~\citep{zhang2016colorful}, context prediction~\citep{nathan2018improvements}, etc. 
The auxiliary tasks in SSL can be constructed using many different mechanisms. Recently, a contrastive mechanism~\citep{hadsell2006dimensionality} has gained increasing attention and demonstrated promising results in several studies~\citep{he2019moco,chen2020mocov2}. The basic idea of contrastive SSL is: generate augmented examples of original data examples, create a predictive task where the goal is to predict whether two augmented examples are from the same original data example or not, and learn the representation network by solving this task. 

\begin{figure}[h]
    \centering
    \includegraphics[width=0.6\textwidth]{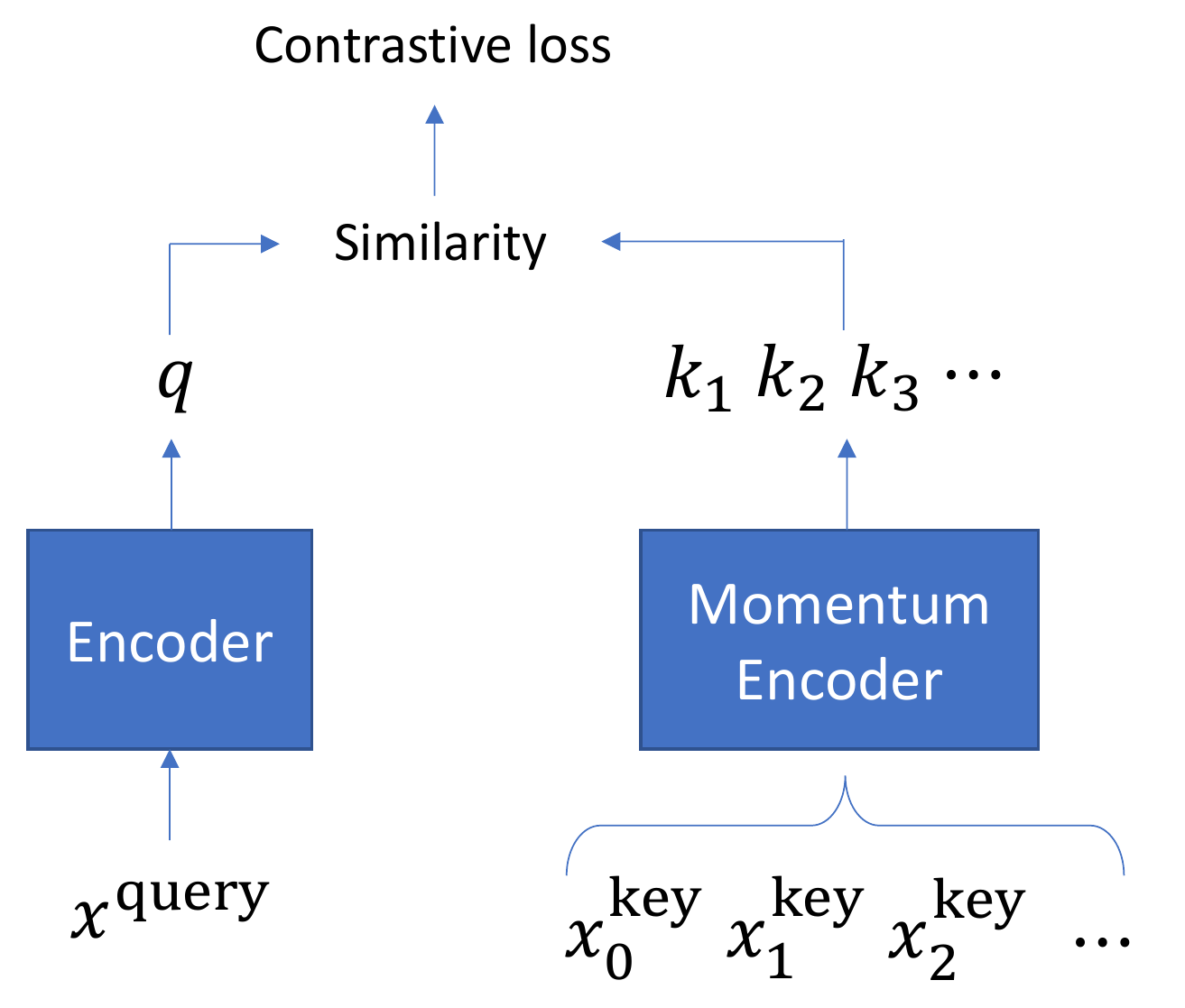}
    \caption{Keys in the queue are encoded using a momentum encoder. Given an augmented data example in the current minibatch (called query) and a key in the queue, they are considered as a positive pair if they originate from the same data example, and a negative pair if otherwise.  A similarity score is calculated between the encoding of the query and the encoding of each key. Contrastive losses are defined on the similarity scores and binary labels.}
    \label{fig:moco}
\end{figure}

Different methods have been proposed to implement contrastive SSL. In SimCLR~\citep{chen2020simple} designed for image data, given the input images, random data augmentation is applied to these images. If two augmented images are created from the same original image, they are labeled as being similar; otherwise, they are labeled as dissimilar. Then SimCLR learns a network to fit these similar/dissimilar binary labels. The network consists of two modules: a feature extraction module $f(\cdot)$ which extracts the latent representation $\mathbf{h}=f(\mathbf{x})$ of an image $\mathbf{x}$  and a multi-layer perceptron $g(\cdot)$ which takes $\mathbf{h}$ as input and generates another latent representation $\mathbf{z}=g(\mathbf{h})$ used for predicting whether two images are similar. Given a similar pair $(\mathbf{x}_i,\mathbf{x}_j)$ and a set of images $\{\mathbf{x}_k\}$ that are dissimilar from $\mathbf{x}_i$, a contrastive loss can be defined as follows:
\begin{equation}
    -\log \frac{\textrm{exp}(\textrm{sim}(\mathbf{z}_i, \mathbf{z}_j)/\tau )}{\textrm{exp}(\textrm{sim}(\mathbf{z}_i, \mathbf{z}_j)/\tau )+\sum_{k}\textrm{exp}(\textrm{sim}(\mathbf{z}_i, \mathbf{z}_k)/\tau )}
\end{equation}
where $\textrm{sim}(\cdot,\cdot)$ denotes cosine similarity between two vectors and $\tau$ is a temperature parameter. SimCLR learns the network weights by minimizing losses of this kind. After training, the feature extraction sub-network is used for downstream tasks and $g(\cdot)$ is discarded.

While SimCLR is easy to implement, it requires a large minibatch size to yield high performance, which is computationally prohibitive. MoCo~\citep{chen2020simple} addresses this problem by using a queue that is independent of minibatch size. This queue contains a dynamic set of  augmented data examples (called keys). In each iteration, the latest minibatch of examples are added into the queue; meanwhile, the oldest minibatch is removed from the queue. In this way, the queue is decoupled from minibatch size. Figure~\ref{fig:moco} shows the architecture of MoCo. The keys are encoded using a momentum encoder. Given an augmented data example (called query) in the current minibatch and a key in the queue, they are considered as a positive pair if they originate from the same image, and a negative pair if otherwise.  A similarity score is calculated between the encoding of the query and the encoding of each key. Contrastive losses are defined on the similarity scores and binary labels.

\section{CERT}
\begin{figure}[h]
    \centering
    \includegraphics[width=0.6\textwidth]{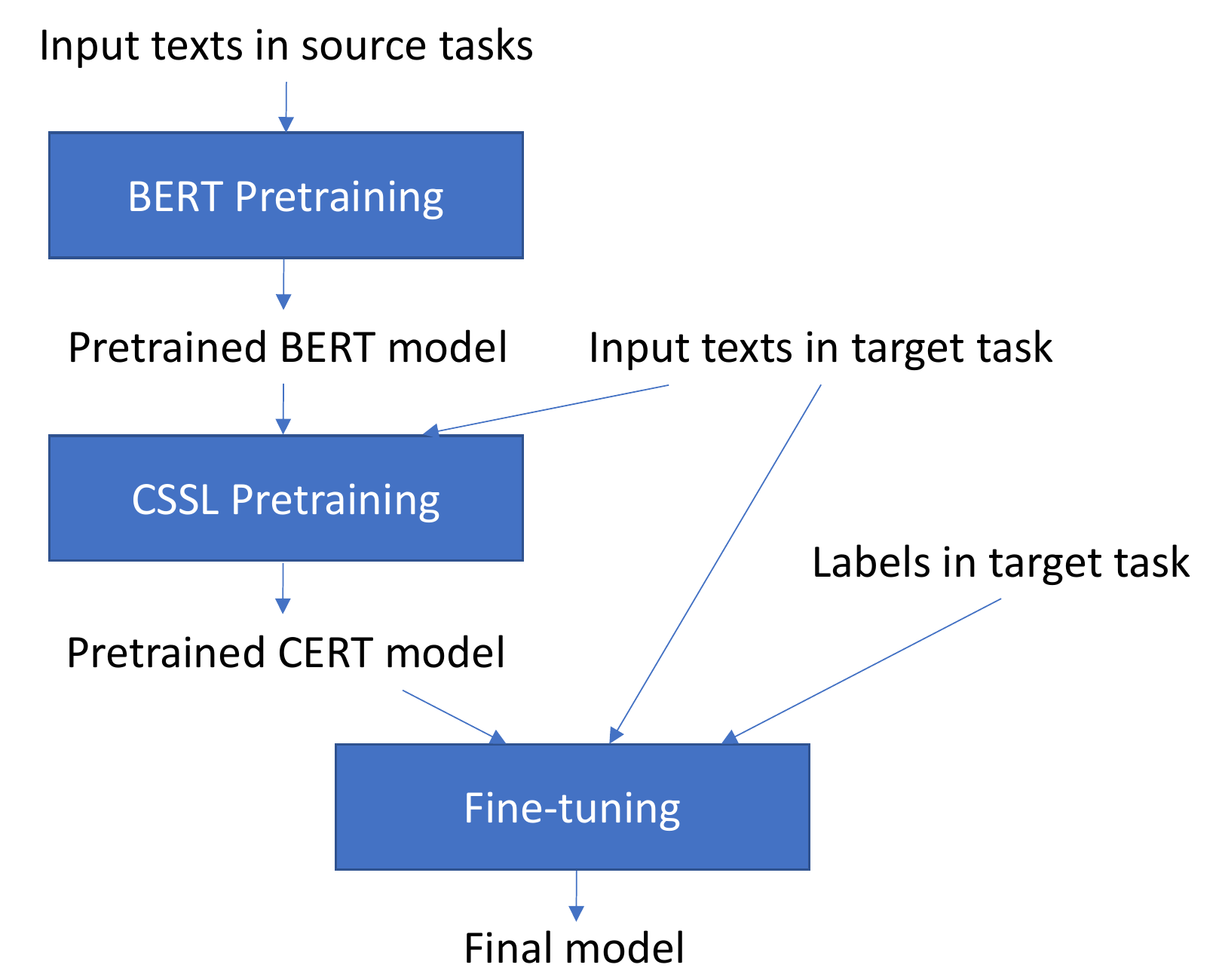}
    \caption{The workflow of CERT. Given the large-scale input texts (without labels) from source tasks, a BERT model is first pretrained  on these texts. Then we continue to train this pretrained BERT model using CSSL on the input texts (without labels) from the target task. We refer to this model as pretrained CERT model. Then we finetune the CERT model using the input texts and their associated labels in the target task and get the final model that performs the target task.}
    \label{fig:cert}
\end{figure}

CERT takes a pretrained language representation model (e.g., BERT) and finetunes it using contrastive self-supervised learning on the input data of the target task. 
Figure~\ref{fig:cert} shows the workflow of CERT. For the ease of presentation, we use BERT as a running example of the pretrained language representation model. But note that CERT can be used on top of other  pretrained language representation models  such as XLNet, T5, etc. as well and is not specific to BERT. 
Given the large-scale input texts (without labels) from source tasks, a BERT model is first pretrained  on these texts. Then we continue to train this pretrained BERT model using CSSL on the input texts (without labels) from the target task. We refer to this model as pretrained CERT model. Then we finetune the CERT model using the input texts and their associated labels in the target task and get the final model that performs the target task.  In CSSL training, CERT augments the original sentences in the target task using back-translation. Two augmented sentences are considered as a positive pair if they are created from the same original sentence and a negative pair  if otherwise. The augmented sentences are encoded with BERT and a similarity score is calculated on the BERT encodings of a pair of sentences. Contrastive losses are defined on the binary labels and similarity scores. Then the pretrained BERT encoder is further finetuned by minimizing the contrastive losses. We use MoCo to implement CSSL to avoid using large minibatches which are computationally heavy.

\begin{figure}[h]
	\begin{center}
 	\includegraphics[width = \columnwidth]{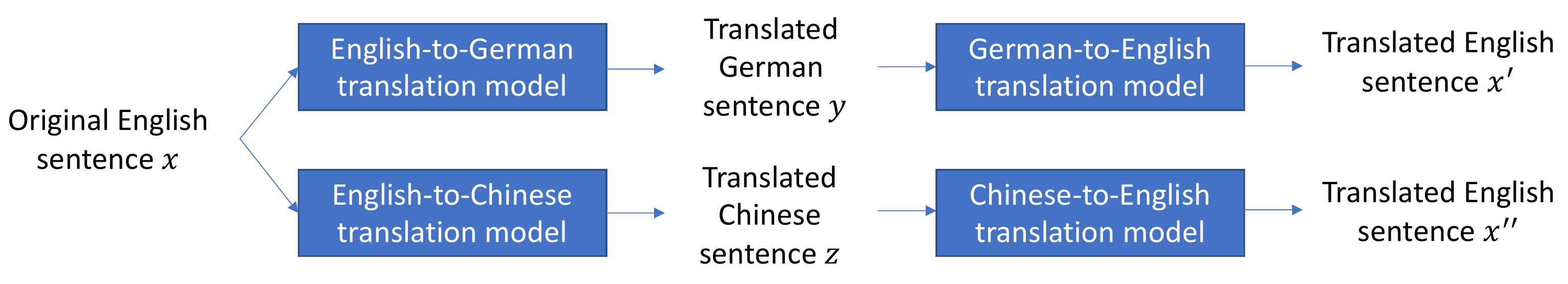}
 	\caption{The workflow of data augmentation based on back translation. 
	}\label{fig:bt}
	\end{center}
 \end{figure}

\paragraph{Data Augmentation}
Figure~\ref{fig:bt} shows the workflow of data augmentation based on back translation.  For each input sentence $x$ in the target task, we augment it using back-translation~\citep{edunov2018understanding}. Without loss of generality, we assume the language in the target task is English. We use an English-to-German machine translation (MT) model to translate $x$ to $y$. Then we use a German-to-English translation model to translate $y$ to $x'$. Then $x'$ is regarded as an augmented sentence of $x$. Similarly, we use an English-to-Chinese MT model and a Chinese-to-English MT model to obtain another augmented sentence $x''$. We use the machine translation models developed in~\citep{britz2017massive} for back-translation.

\paragraph{CSSL Pretraining}
We use MoCo~\citep{he2019moco} to implement CSSL.
Given two augmented sentences, if they originate from the same original sentence, they are labeled as a positive pair; if they are from different sentences, they are labeled as a negative pair. We use a queue to maintain a set of augmented sentences $\{k_i\}_{i=1}^K$ (called keys). Given an augmented sentence $q$ (called query) in the currently sampled minibatch, it is compared against each key in the queue. The query is encoded with a pretrained BERT model $f_q(\cdot;\theta_q)$ where $\theta_q$ denotes the weights. Each key is encoded with a pretrained BERT $f_k(\cdot;\theta_k)$, where the weights $\theta_k$ are updated with momentum:  $\theta_k\gets m \theta_k+(1-m) \theta_q$ ($m$ is the momentum coefficient). Without loss of generality, we assume there is a single key $k_+$ in the queue that forms a positive pair with $q$. A contrastive loss can be defined as:
\begin{equation}
    -\log\frac{\exp(\textrm{sim}(f_q(q;\theta_q), f_k(k_+;\theta_k))/\tau)}{\sum_{i=1}^K\exp(\textrm{sim}(f_q(q;\theta_q), f_k(k_i;\theta_k))/\tau)}
\end{equation}
where $\tau$ is a temperature parameter. The weights of the BERT encoder are finetuned by minimizing losses of such kind. 
Note that in this step, only the input sentences of the target task are used. The labels of these sentences are not touched. To apply the pretrained CERT model to a downstream task, we further finetune its weights on both the input sentences and their labels in the target task.

\section{Experiments}
In this section, we evaluate CERT on eleven natural language understanding tasks in the GLUE~\citep{wang2018glue} benchmark and compare with BERT. 

\subsection{Tasks and Datasets}
The General Language Understanding Evaluation (GLUE) benchmark has 11 tasks, including 2 single-sentence tasks, 3 similarity and paraphrase tasks, and 5 inference tasks. The evaluation metric for a task is accuracy, unless otherwise noted. For each task, labels of the validation set are publicly available while those of the test set are not released. We obtain performance on the test sets by submitting inference results to the GLUE evaluation server\footnote{https://gluebenchmark.com/leaderboard}. 
\begin{itemize}[leftmargin=*]
\item \textbf{CoLA} On the Corpus of Linguistic Acceptability (CoLA)~\citep{warstadt2019neural}, the task is to judge whether a sequence of words is a grammatical English sentence. Matthews correlation coefficient~\citep{matthews1975comparison} is used as the evaluation metric. The higher, the better. 
\item \textbf{SST-2} On the Stanford Sentiment Treebank (SST)~\citep{socher2013recursive}, the task is to judge whether the sentiment of a sentence is positive or negative. 

\item \textbf{MRPC} On the Microsoft Research Paraphrase Corpus (MRPC)~\citep{dolan2005automatically}, the task is to predict whether a pair of sentences are semantically equivalent. Accuracy and F1 are used as evaluation metrics. 

\item \textbf{QQP} On the Quora Question Pairs (QQP)\footnote{https://www.quora.com/q/quoradata/First-Quora-Dataset-Release-Question-Pairs}, the  task is  to determine whether a pair of questions are semantically equivalent. Accuracy and F1 are used as evaluation metrics. 

\item \textbf{STS-B} On the Semantic Textual Similarity Benchmark (STS-B)~\citep{cer2017semeval}, the task is to predict the similarity score (from 1 to 5) of a pair of sentences. Pearson and Spearman correlation coefficients are used for evaluation. 
 
\item \textbf{MNLI} On the Multi-Genre Natural Language Inference (MNLI) corpus~\citep{williams2017broad}, given a premise sentence and a hypothesis sentence, the task is to predict whether the premise entails the hypothesis or not. The task contains two sub-tasks: MNLI-m and MNLI-mm which perform evaluation on matched (in-domain) and mismatched (cross-domain) sections.

\item \textbf{QNLI} On the Stanford Question Answering Dataset~\citep{rajpurkar2016squad}, given a question  and a paragraph, the task is to predict whether the paragraph contains an  answer to the question.

\item \textbf{RTE} The Recognizing Textual Entailment (RTE)~\citep{dagan2005pascal} task is defined as follows: given two sentences, judge whether one sentence can entail another. Entailment means the truth of one sentence follows from the other sentence.  

\item \textbf{WNLI} In the Winograd Schema Challenge~\citep{levesque2012winograd}, the task is to read a sentence with a pronoun and select the reference of that pronoun from
a list of choices.

\end{itemize}
Table~\ref{tab:data-stats} shows the statistics of data split in each task.

\begin{table}[h]
    \centering
    \begin{tabular}{l|c|c|c|c|c|c|c|c|c}
    \hline
       & COLA & RTE & QNLI & STS-B  & MRPC & WNLI & SST-2 & \shortstack[c]{MNLI\\(m/mm)} & QQP \\
        \hline
Train & 8551 & 2491 & 104742 & 5749 & 3668 & 636 & 67350 & 392702 & 363871\\
Dev & 1043&  278 & 5462 & 1500 & 408 & 72 & 873 & 9815/9832 & 40432\\
Test & 1064 & 2985 & 5462 & 1379 & 1725 & 147 & 1821 & 9796/9847 & 390965\\
    \hline
    \end{tabular}
    \caption{Dataset statistics.}
    \label{tab:data-stats}
\end{table}

\subsection{Experimental Settings}
In MoCo, the size of the queue was set to 96606. The coefficient of MoCo momentum of updating the key encoder was set to 0.999. The temperature parameter in contrastive loss was set to 0.07. Multi-layer perceptron head was used.
For MoCo training, a stochastic gradient descent solver with momentum was used. The minibatch size was set to 16. The initial learning rate was set to 4e-5. The learning rate was adjusted using  cosine scheduling. The number of epochs was set to 100.   Weight decay was used with a coefficient of 1e-5. For finetuning on GLUE tasks, the maximum sequence length was set to 128. Minibatch size was set to 16. The learning rate was set to 3e-5 for CoLA, MNLI, STS-B; 2e-5 for RTE, QNLI, MRPC, SST-2, WNLI; and 1e-5 for QQP. The number of training epochs was set to 20 for CoLA; 5 for RTE, WNLI, QQP; 3 for QNLI, MNLI; 15 for MRPC; 4 for SST-2; and 10 for STS-B.

\subsection{Results}
Table~\ref{tab:dev} shows the performance on the development sets in GLUE tasks. Following~\citep{lan2019albert}, we perform 5 random restarts of the finetuning and report the median and best performance among the 5 runs. Due to randomness of the 5 restarts, our median performance is not the same as that reported in~\citep{lan2019albert}. But generally they are close. From this table, we make the following observations. First, in terms of median performance, our proposed CERT outperforms BERT (our run) on 5 tasks, including CoLA, RTE, QNLI, SST-2, and QQP. 
On the other 5 tasks, CERT is on par with BERT. This demonstrates the effectiveness of CERT in learning better language representations via contrastive self-supervised learning.  Second, in terms of best performance, CERT outperforms BERT on 7 tasks, including CoLA, RTE, QNLI, STS-B, MNLI-m, MNLI-mm, and QQP. CERT is on par with BERT on MRPC and WNLI. CERT performs less well than BERT on SST-2. These results further demonstrate the effectiveness of CERT. Third, the improvement of CERT over BERT is more significant on tasks where the training data is small, such as CoLA and RTE. One possible reason is that small-sized training data is more prone to overfitting, which makes the necessity of CSSL pretraining more prominent. Fourth, on small-sized datasets such as CoLA and RTE, CERT achieves great improvement over BERT. This shows that CERT is promising in solving low-resource NLP tasks where the amount of training data is limited. 

\begin{table}[h]
    \centering
    \small
    \begin{tabular}{p{2.1cm}|c|c|c|c|c|c|c|c|c}
    \hline
       & CoLA & RTE & QNLI & STS-B  & MRPC & WNLI & SST-2 &   \shortstack[c]{MNLI\\(m/mm)} & QQP \\
        \hline
\shortstack[l]{BERT \\(reported in\\ Lan et al.2019\\
 \\ median)} & 60.6& 70.4& 92.3& 90.0 & - & -  & 93.2 & 86.6/-  & 91.3\\
\hline
\shortstack[l]{ BERT\\ (our run, median)} & 59.8&  71.4 & 91.9 & 89.8 & 91.1  & 56.3 & 93.4 &86.3/86.2 & 91.2\\
 \hline
\shortstack[l]{ CERT\\ (ours, median)} &  62.1& 72.2& 92.3 & 89.8 & 91.0 & 56.3 & 93.6 & 86.3/86.2& 91.4 \\
\hline
\hline
 \shortstack[l]{ BERT \\(our run, best)} & 60.9 & 72.1 & 92.1 & 90.1 & 92.5 & 56.3 & 94.4 & 86.4/86.3& 91.4\\
  \hline
\shortstack[l]{    CERT \\(ours, best)} & 62.9 &74.0 & 92.5 & 90.6  & 92.5 & 56.3 & 93.9 & 86.6/86.5 & 91.7 \\
\hline
\hline
XLNet   & 63.6&83.8 & 93.9 & 91.8 & - & - & 95.6 & 89.8/- & 91.8\\
ERNIE2   & 65.4& 85.2& 94.3 & 92.3 & - & - & 96.0 & 89.1/- & 92.5 \\
RoBERTa   &68.0 &86.6 & 94.7 &92.4 & - & - & 96.4 &  90.2/- & 92.2\\
ALBERT   &71.4 &89.2 & 95.3 &93.0 & - & - & 96.9 & 90.8/- & 92.2\\
    \hline
    \end{tabular}
    \caption{Performance on the validation dataset. The metric for MRPC is F1. The metric for QQP is accuracy. The metric for STS-B is Pearson correlation.}
    \label{tab:dev}
\end{table}

\begin{table}[h]
    \centering
    \small
    \begin{tabular}{l|c|c||c|c|c|c|c}
    \hline
    & BERT & CERT & XLNet & ERNIE2& T5 & RoBERTa & ALBERT \\
    \hline 
    Average &80.5 &80.7 &- &90.4 &90.3 &88.1 &- \\
    \hline
    
    CoLA & 60.5& 58.9& 70.2& 74.4& 71.6& 67.8&69.1\\
    
    RTE & 70.1& 71.2& 88.5& 90.9& 92.8& 88.2&89.2\\
    
    QNLI & 92.7& 93.0& -& 96.6& 96.9& 95.4&-\\
    
    STS-B & 87.6/86.5& 87.9/86.8& 93.0/92.6& 93.0/92.6& 93.1/92.8& 92.2/91.9 &92.5/92.0\\    
    
    MRPC & 89.3/85.4& 89.8/85.9& 92.9/90.5& 93.5/91.4& 92.8/90.4& 92.3/89.8& 93.4/91.2\\
    
    WNLI & 65.1& 65.1& 92.5& 94.5& 94.5& 89.0&91.8 \\    
   
    SST-2 & 94.9& 94.6& 97.1& 97.5& 97.5& 96.7 &97.1\\

    MNLI-m & 86.7& 87.2& 90.9& 91.4& 92.2& 90.8 &91.3\\

    MNLI-mm & 85.9& 86.4& 90.9& 91.0& 91.9& 90.2 &91.0\\

    QQP & 72.1/89.3& 72.5/90.3& 74.7/90.4& 75.2/90.9& 75.1/90.6& 74.3/90.2& 74.2/90.5\\

    AX & 39.6& 39.6& 48.4& 51.7& 53.1& 48.7& 50.2\\
    \hline
    \end{tabular}
    \caption{Performance on the test datasets of GLUE. The metrics for MRPC and QQP are F1/accuracy. The metrics for STS-B are Pearson correlation and Spearman correlation.}
    \label{tab:da-ablation}
\end{table}



For the convenience of the readers, we also show the results of state-of-the-art pretraining methods including XLNet~\citep{yang2019xlnet}, RoBERTa~\citep{liu2019roberta}, ERNIE 2.0~\citep{sun2019ernie}, and ALBERT~\citep{lan2019albert}. CERT achieves a performance close to XLNet on CoLA, with a much lower consumption of computing resources and on a much smaller text corpus. XLNet, RoBERTa, ERNIE 2.0, and ALBERT are trained on hundreds of to thousands of GPU machines for several days, while CERT is trained using a single GPU for a dozen of hours. In addition, these models are trained on tens of or hundreds of gigabytes of texts while CERT is only trained on about 400 KB of texts. CERT can be built on top of these pretrained models as well, which will be left for future study.

Table~\ref{tab:da-ablation} shows the performance on the test datasets of GLUE. Among the 11 tasks, CERT outperforms BERT on 7 tasks, including RTE, QNLI, STS-B, MRPC, MNLI-m, MNLI-mm, QQP. CERT achieves the same performance as BERT on WNLI and AX. 
CERT performs worse than BERT on CoLA and SST-2. CERT achieves an average score of 80.7, which is better than BERT. Overall, CERT performs better than BERT, which further demonstrates that contrastive self-supervised learning is an effective approach for learning better representations of language. While CERT achieves much better performance than BERT on the validation set of CoLA, it performs worse than BERT on the CoLA test set. This is probably because the test set and validation set in CoLA have a large domain difference. Table~\ref{tab:da-ablation} also lists the performance of other state-of-the-art methods. For the next step, we plan to replace the BERT-based sentence encoder in CERT with XLNet, ERNIE2, T5, RoBERTa, and ALBERT, to see whether CSSL pretraining can improve the performance of these encoders.

\paragraph{Ablation on Data Augmentation} One key ingredient in CERT is to create augmented sentences. By default, we use back-translation for augmentation. It is interesting to investigate other augmentation methods as well. We compare back-translation with a recently proposed text augmentation method -- Easy Data Augmentation (EDA) \citep{wei2019eda}. Given a sentence in the training set, EDA randomly chooses and performs one of the following operations: synonym replacement, random insertion, random swap, and random deletion. 

\begin{table}[h]
    \centering
    \begin{tabular}{l|c|c}
    \hline
         & CoLA
           & STS-B \\
        \hline
 Back-translation (median) & 62.1 & 89.8\\
 EDA (median) &  60.5 & 89.9\\
 \hline
  Back-translation (best) &  62.9 & 90.6\\
 EDA (best) &  62.3 & 90.2\\
    \hline
    \end{tabular}
    \caption{Performance on CoLA and STS-B, under different data-augmentation methods.}
    \label{tab:da-ab}
\end{table} 
Table~\ref{tab:da-ab} shows the results achieved by CERT using back-translation and EDA for augmentation respectively, on the CoLA and RTE tasks. As can be seen, in general, back-translation achieves better results than EDA, except that the median performance of back-translation on STS-B is 0.1\% lower than that of EDA. The reason that back-translation works better is probably because back-translation performs augmentation at the sentence level globally: the entire sentence is translated back and forth, while EDA performs augmentation at the word/phrase level locally. Therefore, back-translation can better capture the global semantics of sentences while EDA captures local semantics. Another reason might be that the sentences augmented by back-translation are more different from the original ones. In contrast, the sentences augmented by EDA are close to the original ones since EDA performs local edits of the original sentences and keeps the majority of a sentence untouched. It is more difficult to predict whether two augmented sentences by back-translation are from the same original sentence than to predict those augmented by EDA. Solving a more challenging CSSL task usually leads to better representations. 

\begin{table}[h]
    \centering
    \begin{tabular}{l|c|c }
    \hline
         & CoLA & STS-B\\
        \hline
 Target Training Data (median) & 62.1 &   89.8 \\
 BERT Training Data (median) &  60.0 & 90.3 \\
 \hline
  Target Training Data (best) &  62.9 &  90.6 \\
 BERT Training Data (best) &  61.3 &  90.8\\
    \hline
    \end{tabular}
    \caption{Performance on CoLA and STS-B, under different pretraining corpus of CSSL.}
    \label{tab:pre-co}
\end{table}

\paragraph{Ablation on CSSL Pretraining Corpora} Another interesting point to study is what corpora should be used for training CSSL. In CERT, by default, we use the training data (excluding labels) of the target task for CSSL. We compare with the following setting. We randomly sample $x$ sentences from the training corpora of BERT, where $x$ is the number of training sentences in the target task. Table~\ref{tab:pre-co} shows the performance on CoLA and STS-B, with CSSL pretrained on different corpora. On the CoLA task, better performance is achieved when using the training data of CoLA for CSSL. On the STS-B task, better performance is achieved when using the BERT pretraining corpus for CSSL. From this study, we do not reach a clear conclusion regarding which one is better. In practice, it might be useful to try both and find out which one is more effective. 


\section{Related Works}
\subsection{Pretraining for learning language representation}
Recently, pretraining on large-scale text corpus for language representation learning has achieved substantial success.  The GPT model~\citep{radford2018improving} is a language model (LM) based on Transformer. Different from Transformer which defines a conditional probability on an output sequence given an input sequence, GPT defines a marginal probability on a single sequence. In GPT, the conditional probability of the next token given the historical sequence is defined using the Transformer decoder. The weight parameters are learned by maximizing the likelihood on the sequence of tokens.  
GPT-2~\citep{radford2019language} is an extension of GPT, which modifies GPT by moving layer normalization  to the input of each sub-block and adding an
additional layer normalization  after the final self-attention block. Byte pair encoding (BPE) \citep{sennrich2015neural} is used to represent the input sequence of tokens. BERT-GPT~\citep{wu2019importance} is a model used for sequence-to-sequence modeling where pretrained BERT is used to encode the input text and GPT is used to generate the output text.
In BERT-GPT, the pretraining of the BERT encoder and the GPT decoder is conducted separately, which may lead to inferior performance.  Auto-Regressive
Transformers (BART)~\citep{lewis2019bart} has a similar architecture as BERT-GPT, but trains the BERT encoder and GPT decoder jointly. To pretrain the BART weights, the input text is corrupted randomly, such as token masking, token deletion, text infilling, etc., then the network is
learned to reconstruct the original text.  ALBERT~\citep{lan2019albert} uses parameter-reduction methods to reduce the memory consumption and increase the training speed of BERT. It also introduces a self-supervised loss which models inter-sentence coherence. RoBERTa~\citep{liu2019roberta} is a replication study of BERT pretraining. It shows that the performance of BERT can be significantly improved by carefully tuning the training process, such as (1) training the model longer, with bigger batches, over more data; (2) removing the next sentence prediction objective; (3) training on longer sequences, etc. XLNet~\citep{yang2019xlnet} learns bidirectional contexts by maximizing the expected likelihood over all permutations of the factorization order and uses a generalized autoregressive pretraining mechanism to overcome the pretrain-finetune discrepancy of BERT. T5~\citep{raffel2019exploring} compared pretraining objectives, architectures, unlabeled datasets, transfer approaches on a wide range of language understanding tasks and proposed a unified framework that casts these tasks as a text-to-text task. The unified model was trained on a large Colossal Clean Crawled Corpus, which was then transferred to diverse downstream tasks. ERNIE 2.0~\citep{sun2019ernie} proposed a continual pretraining framework which builds and learns incrementally pretraining tasks through constant multi-task learning, to capture the lexical, syntactic and semantic information from training corpora.

\subsection{Contrastive Self-supervised learning}
Contrastive self-supervised learning has arisen much research interest recently. 
Hénaff et al.~\citep{henaff2019data} studied data-efficient image recognition based on  contrastive predictive coding~\citep{oord2018representation}, which  predicts the future in latent space by using powerful autoregressive models. Srinivas et al.~\citep{srinivas2020curl} proposed to learn  contrastive unsupervised representations for reinforcement learning. Khosla et al.~\citep{khosla2020supervised} investigated supervised contrastive learning, where clusters of points belonging to the same class are pulled together in embedding space, while clusters of samples from different classes are pushed apart. Klein and  Nabi~\citep{klein2020contrastive} proposed a contrastive self-supervised learning approach for commonsense reasoning. He et al.~\citep{he2020sample} proposed an  Self-Trans approach which applies contrastive self-supervised learning on top of  networks pretrained by transfer learning. 

\section{Conclusions and Future Works}
In this work, we propose CERT, a pretraining approach for language representation learning. CERT takes a pretrained language representation model such as BERT and continues to train it using contrastive self-supervised learning on the input texts of the target task. It uses back-translation to generate augmented sentences for each original sentence in the target data, and trains the network by predicting whether two augments are from the same original sentence or not. Then the pretrained CERT model is finetuned on the input texts and their labels in the target task. We evaluate CERT on 11 natural language understanding tasks in the GLUE benchmark. On both test set and validation set, CERT outperforms BERT on majority of tasks, which demonstrates the effectiveness of contrastive self-supervised learning in learning language representations.  For future works, we plan to study more challenging loss functions for self-supervised learning. We are interested in investigating a ranking-based loss, where each sentence is augmented with a ranked list of sentences which have decreasing discrepancy with the original sentence. The auxiliary task is to predict the order given the augmented sentences. Predicting an order is presumably more challenging than binary classification as in contrastive SSL and may facilitate the learning of better representations.

\bibliography{release}

\begin{thebibliography}{45}
\providecommand{\natexlab}[1]{#1}
\providecommand{\url}[1]{\texttt{#1}}
\expandafter\ifx\csname urlstyle\endcsname\relax
  \providecommand{\doi}[1]{doi: #1}\else
  \providecommand{\doi}{doi: \begingroup \urlstyle{rm}\Url}\fi

\bibitem[Ba et~al.(2016)Ba, Kiros, and Hinton]{ba2016layer}
Jimmy~Lei Ba, Jamie~Ryan Kiros, and Geoffrey~E Hinton.
\newblock Layer normalization.
\newblock \emph{arXiv preprint arXiv:1607.06450}, 2016.

\bibitem[Britz et~al.(2017)Britz, Goldie, Luong, and Le]{britz2017massive}
Denny Britz, Anna Goldie, Minh-Thang Luong, and Quoc Le.
\newblock Massive exploration of neural machine translation architectures.
\newblock \emph{arXiv preprint arXiv:1703.03906}, 2017.

\bibitem[Cer et~al.(2017)Cer, Diab, Agirre, Lopez-Gazpio, and
  Specia]{cer2017semeval}
Daniel Cer, Mona Diab, Eneko Agirre, Inigo Lopez-Gazpio, and Lucia Specia.
\newblock Semeval-2017 task 1: Semantic textual similarity-multilingual and
  cross-lingual focused evaluation.
\newblock \emph{arXiv preprint arXiv:1708.00055}, 2017.

\bibitem[Chen et~al.(2020{\natexlab{a}})Chen, Kornblith, Norouzi, and
  Hinton]{chen2020simple}
Ting Chen, Simon Kornblith, Mohammad Norouzi, and Geoffrey Hinton.
\newblock A simple framework for contrastive learning of visual
  representations.
\newblock \emph{arXiv preprint arXiv:2002.05709}, 2020{\natexlab{a}}.

\bibitem[Chen et~al.(2020{\natexlab{b}})Chen, Fan, Girshick, and
  He]{chen2020mocov2}
Xinlei Chen, Haoqi Fan, Ross Girshick, and Kaiming He.
\newblock Improved baselines with momentum contrastive learning.
\newblock \emph{arXiv preprint arXiv:2003.04297}, 2020{\natexlab{b}}.

\bibitem[Chung et~al.(2014)Chung, Gulcehre, Cho, and
  Bengio]{chung2014empirical}
Junyoung Chung, Caglar Gulcehre, KyungHyun Cho, and Yoshua Bengio.
\newblock Empirical evaluation of gated recurrent neural networks on sequence
  modeling.
\newblock \emph{arXiv preprint arXiv:1412.3555}, 2014.

\bibitem[Dagan et~al.(2005)Dagan, Glickman, and Magnini]{dagan2005pascal}
Ido Dagan, Oren Glickman, and Bernardo Magnini.
\newblock The pascal recognising textual entailment challenge.
\newblock In \emph{Machine Learning Challenges Workshop}, pages 177--190.
  Springer, 2005.

\bibitem[Devlin et~al.(2018)Devlin, Chang, Lee, and Toutanova]{devlin2018bert}
Jacob Devlin, Ming-Wei Chang, Kenton Lee, and Kristina Toutanova.
\newblock Bert: Pre-training of deep bidirectional transformers for language
  understanding.
\newblock \emph{arXiv preprint arXiv:1810.04805}, 2018.

\bibitem[Dolan and Brockett(2005)]{dolan2005automatically}
William~B Dolan and Chris Brockett.
\newblock Automatically constructing a corpus of sentential paraphrases.
\newblock In \emph{Proceedings of the Third International Workshop on
  Paraphrasing (IWP2005)}, 2005.

\bibitem[Edunov et~al.(2018)Edunov, Ott, Auli, and
  Grangier]{edunov2018understanding}
Sergey Edunov, Myle Ott, Michael Auli, and David Grangier.
\newblock Understanding back-translation at scale.
\newblock \emph{arXiv preprint arXiv:1808.09381}, 2018.

\bibitem[Gidaris et~al.(2018)Gidaris, Singh, and
  Komodakis]{gidaris2018unsupervised}
Spyros Gidaris, Praveer Singh, and Nikos Komodakis.
\newblock Unsupervised representation learning by predicting image rotations.
\newblock \emph{arXiv preprint arXiv:1803.07728}, 2018.

\bibitem[Hadsell et~al.(2006)Hadsell, Chopra, and
  LeCun]{hadsell2006dimensionality}
Raia Hadsell, Sumit Chopra, and Yann LeCun.
\newblock Dimensionality reduction by learning an invariant mapping.
\newblock In \emph{2006 IEEE Computer Society Conference on Computer Vision and
  Pattern Recognition (CVPR'06)}, volume~2, pages 1735--1742. IEEE, 2006.

\bibitem[He et~al.(2016)He, Zhang, Ren, and Sun]{he2016deep}
Kaiming He, Xiangyu Zhang, Shaoqing Ren, and Jian Sun.
\newblock Deep residual learning for image recognition.
\newblock In \emph{Proceedings of the IEEE conference on computer vision and
  pattern recognition}, pages 770--778, 2016.

\bibitem[He et~al.(2019)He, Fan, Wu, Xie, and Girshick]{he2019moco}
Kaiming He, Haoqi Fan, Yuxin Wu, Saining Xie, and Ross Girshick.
\newblock Momentum contrast for unsupervised visual representation learning.
\newblock \emph{arXiv preprint arXiv:1911.05722}, 2019.

\bibitem[He et~al.(2020)He, Yang, Zhang, Zhao, Zhang, Xing, and
  Xie]{he2020sample}
Xuehai He, Xingyi Yang, Shanghang Zhang, Jinyu Zhao, Yichen Zhang, Eric Xing,
  and Pengtao Xie.
\newblock Sample-efficient deep learning for covid-19 diagnosis based on ct
  scans.
\newblock \emph{medRxiv}, 2020.

\bibitem[H{\'e}naff et~al.(2019)H{\'e}naff, Razavi, Doersch, Eslami, and
  Oord]{henaff2019data}
Olivier~J H{\'e}naff, Ali Razavi, Carl Doersch, SM~Eslami, and Aaron van~den
  Oord.
\newblock Data-efficient image recognition with contrastive predictive coding.
\newblock \emph{arXiv preprint arXiv:1905.09272}, 2019.

\bibitem[Hochreiter and Schmidhuber(1997)]{hochreiter1997long}
Sepp Hochreiter and J{\"u}rgen Schmidhuber.
\newblock Long short-term memory.
\newblock \emph{Neural computation}, 9\penalty0 (8):\penalty0 1735--1780, 1997.

\bibitem[Khosla et~al.(2020)Khosla, Teterwak, Wang, Sarna, Tian, Isola,
  Maschinot, Liu, and Krishnan]{khosla2020supervised}
Prannay Khosla, Piotr Teterwak, Chen Wang, Aaron Sarna, Yonglong Tian, Phillip
  Isola, Aaron Maschinot, Ce~Liu, and Dilip Krishnan.
\newblock Supervised contrastive learning.
\newblock \emph{arXiv preprint arXiv:2004.11362}, 2020.

\bibitem[Klein and Nabi(2020)]{klein2020contrastive}
Tassilo Klein and Moin Nabi.
\newblock Contrastive self-supervised learning for commonsense reasoning.
\newblock \emph{arXiv preprint arXiv:2005.00669}, 2020.

\bibitem[Lan et~al.(2019)Lan, Chen, Goodman, Gimpel, Sharma, and
  Soricut]{lan2019albert}
Zhenzhong Lan, Mingda Chen, Sebastian Goodman, Kevin Gimpel, Piyush Sharma, and
  Radu Soricut.
\newblock Albert: A lite bert for self-supervised learning of language
  representations.
\newblock \emph{arXiv preprint arXiv:1909.11942}, 2019.

\bibitem[Levesque et~al.(2012)Levesque, Davis, and
  Morgenstern]{levesque2012winograd}
Hector Levesque, Ernest Davis, and Leora Morgenstern.
\newblock The winograd schema challenge.
\newblock In \emph{Thirteenth International Conference on the Principles of
  Knowledge Representation and Reasoning}, 2012.

\bibitem[Lewis et~al.(2019)Lewis, Liu, Goyal, Ghazvininejad, Mohamed, Levy,
  Stoyanov, and Zettlemoyer]{lewis2019bart}
Mike Lewis, Yinhan Liu, Naman Goyal, Marjan Ghazvininejad, Abdelrahman Mohamed,
  Omer Levy, Ves Stoyanov, and Luke Zettlemoyer.
\newblock Bart: Denoising sequence-to-sequence pre-training for natural
  language generation, translation, and comprehension.
\newblock \emph{arXiv preprint arXiv:1910.13461}, 2019.

\bibitem[Liu et~al.(2019)Liu, Ott, Goyal, Du, Joshi, Chen, Levy, Lewis,
  Zettlemoyer, and Stoyanov]{liu2019roberta}
Yinhan Liu, Myle Ott, Naman Goyal, Jingfei Du, Mandar Joshi, Danqi Chen, Omer
  Levy, Mike Lewis, Luke Zettlemoyer, and Veselin Stoyanov.
\newblock Roberta: A robustly optimized bert pretraining approach.
\newblock \emph{arXiv preprint arXiv:1907.11692}, 2019.

\bibitem[Matthews(1975)]{matthews1975comparison}
Brian~W Matthews.
\newblock Comparison of the predicted and observed secondary structure of t4
  phage lysozyme.
\newblock \emph{Biochimica et Biophysica Acta (BBA)-Protein Structure},
  405\penalty0 (2):\penalty0 442--451, 1975.

\bibitem[Nathan~Mundhenk et~al.(2018)Nathan~Mundhenk, Ho, and
  Chen]{nathan2018improvements}
T~Nathan~Mundhenk, Daniel Ho, and Barry~Y Chen.
\newblock Improvements to context based self-supervised learning.
\newblock In \emph{Proceedings of the IEEE Conference on Computer Vision and
  Pattern Recognition}, pages 9339--9348, 2018.

\bibitem[Oord et~al.(2018)Oord, Li, and Vinyals]{oord2018representation}
Aaron van~den Oord, Yazhe Li, and Oriol Vinyals.
\newblock Representation learning with contrastive predictive coding.
\newblock \emph{arXiv preprint arXiv:1807.03748}, 2018.

\bibitem[Pathak et~al.(2016)Pathak, Krahenbuhl, Donahue, Darrell, and
  Efros]{pathak2016context}
Deepak Pathak, Philipp Krahenbuhl, Jeff Donahue, Trevor Darrell, and Alexei~A
  Efros.
\newblock Context encoders: Feature learning by inpainting.
\newblock In \emph{Proceedings of the IEEE conference on computer vision and
  pattern recognition}, pages 2536--2544, 2016.

\bibitem[Radford et~al.({\natexlab{a}})Radford, Narasimhan, Salimans, and
  Sutskever]{radford2018improving}
Alec Radford, Karthik Narasimhan, Tim Salimans, and Ilya Sutskever.
\newblock Improving language understanding by generative pre-training.
\newblock {\natexlab{a}}.

\bibitem[Radford et~al.({\natexlab{b}})Radford, Wu, Child, Luan, Amodei, and
  Sutskever]{radford2019language}
Alec Radford, Jeffrey Wu, Rewon Child, David Luan, Dario Amodei, and Ilya
  Sutskever.
\newblock Language models are unsupervised multitask learners.
\newblock {\natexlab{b}}.

\bibitem[Raffel et~al.(2019)Raffel, Shazeer, Roberts, Lee, Narang, Matena,
  Zhou, Li, and Liu]{raffel2019exploring}
Colin Raffel, Noam Shazeer, Adam Roberts, Katherine Lee, Sharan Narang, Michael
  Matena, Yanqi Zhou, Wei Li, and Peter~J Liu.
\newblock Exploring the limits of transfer learning with a unified text-to-text
  transformer.
\newblock \emph{arXiv preprint arXiv:1910.10683}, 2019.

\bibitem[Rajpurkar et~al.(2016)Rajpurkar, Zhang, Lopyrev, and
  Liang]{rajpurkar2016squad}
Pranav Rajpurkar, Jian Zhang, Konstantin Lopyrev, and Percy Liang.
\newblock Squad: 100,000+ questions for machine comprehension of text.
\newblock \emph{arXiv preprint arXiv:1606.05250}, 2016.

\bibitem[Sennrich et~al.(2015)Sennrich, Haddow, and Birch]{sennrich2015neural}
Rico Sennrich, Barry Haddow, and Alexandra Birch.
\newblock Neural machine translation of rare words with subword units.
\newblock \emph{arXiv preprint arXiv:1508.07909}, 2015.

\bibitem[Socher et~al.(2013)Socher, Perelygin, Wu, Chuang, Manning, Ng, and
  Potts]{socher2013recursive}
Richard Socher, Alex Perelygin, Jean Wu, Jason Chuang, Christopher~D Manning,
  Andrew~Y Ng, and Christopher Potts.
\newblock Recursive deep models for semantic compositionality over a sentiment
  treebank.
\newblock In \emph{Proceedings of the 2013 conference on empirical methods in
  natural language processing}, pages 1631--1642, 2013.

\bibitem[Srinivas et~al.(2020)Srinivas, Laskin, and Abbeel]{srinivas2020curl}
Aravind Srinivas, Michael Laskin, and Pieter Abbeel.
\newblock Curl: Contrastive unsupervised representations for reinforcement
  learning.
\newblock \emph{arXiv preprint arXiv:2004.04136}, 2020.

\bibitem[Sun et~al.(2019)Sun, Wang, Li, Feng, Tian, Wu, and Wang]{sun2019ernie}
Yu~Sun, Shuohuan Wang, Yukun Li, Shikun Feng, Hao Tian, Hua Wu, and Haifeng
  Wang.
\newblock Ernie 2.0: A continual pre-training framework for language
  understanding.
\newblock \emph{arXiv preprint arXiv:1907.12412}, 2019.

\bibitem[Sutskever et~al.(2014)Sutskever, Vinyals, and
  Le]{sutskever2014sequence}
Ilya Sutskever, Oriol Vinyals, and Quoc~V Le.
\newblock Sequence to sequence learning with neural networks.
\newblock In \emph{Advances in neural information processing systems}, pages
  3104--3112, 2014.

\bibitem[Vaswani et~al.(2017)Vaswani, Shazeer, Parmar, Uszkoreit, Jones, Gomez,
  Kaiser, and Polosukhin]{vaswani2017attention}
Ashish Vaswani, Noam Shazeer, Niki Parmar, Jakob Uszkoreit, Llion Jones,
  Aidan~N Gomez, {\L}ukasz Kaiser, and Illia Polosukhin.
\newblock Attention is all you need.
\newblock In \emph{Advances in neural information processing systems}, pages
  5998--6008, 2017.

\bibitem[Wang et~al.(2018)Wang, Singh, Michael, Hill, Levy, and
  Bowman]{wang2018glue}
Alex Wang, Amanpreet Singh, Julian Michael, Felix Hill, Omer Levy, and Samuel~R
  Bowman.
\newblock Glue: A multi-task benchmark and analysis platform for natural
  language understanding.
\newblock \emph{arXiv preprint arXiv:1804.07461}, 2018.

\bibitem[Warstadt et~al.(2019)Warstadt, Singh, and Bowman]{warstadt2019neural}
Alex Warstadt, Amanpreet Singh, and Samuel~R Bowman.
\newblock Neural network acceptability judgments.
\newblock \emph{Transactions of the Association for Computational Linguistics},
  7:\penalty0 625--641, 2019.

\bibitem[Wei and Zou(2019)]{wei2019eda}
Jason~W Wei and Kai Zou.
\newblock Eda: Easy data augmentation techniques for boosting performance on
  text classification tasks.
\newblock \emph{arXiv preprint arXiv:1901.11196}, 2019.

\bibitem[Williams et~al.(2017)Williams, Nangia, and Bowman]{williams2017broad}
Adina Williams, Nikita Nangia, and Samuel~R Bowman.
\newblock A broad-coverage challenge corpus for sentence understanding through
  inference.
\newblock \emph{arXiv preprint arXiv:1704.05426}, 2017.

\bibitem[Wu et~al.(2019)Wu, Li, Zhou, Zeng, and Yu]{wu2019importance}
Qingyang Wu, Lei Li, Hao Zhou, Ying Zeng, and Zhou Yu.
\newblock Importance-aware learning for neural headline editing.
\newblock \emph{arXiv preprint arXiv:1912.01114}, 2019.

\bibitem[Wu et~al.(2018)Wu, Xiong, Yu, and Lin]{wu2018unsupervised}
Zhirong Wu, Yuanjun Xiong, Stella~X Yu, and Dahua Lin.
\newblock Unsupervised feature learning via non-parametric instance
  discrimination.
\newblock In \emph{Proceedings of the IEEE Conference on Computer Vision and
  Pattern Recognition}, pages 3733--3742, 2018.

\bibitem[Yang et~al.(2019)Yang, Dai, Yang, Carbonell, Salakhutdinov, and
  Le]{yang2019xlnet}
Zhilin Yang, Zihang Dai, Yiming Yang, Jaime Carbonell, Russ~R Salakhutdinov,
  and Quoc~V Le.
\newblock Xlnet: Generalized autoregressive pretraining for language
  understanding.
\newblock In \emph{Advances in neural information processing systems}, pages
  5754--5764, 2019.

\bibitem[Zhang et~al.(2016)Zhang, Isola, and Efros]{zhang2016colorful}
Richard Zhang, Phillip Isola, and Alexei~A Efros.
\newblock Colorful image colorization.
\newblock In \emph{European conference on computer vision}, pages 649--666.
  Springer, 2016.

\end{thebibliography}

\end{document}